\documentclass{article} 
\usepackage{proceedings}
\usepackage{latexsym}
\usepackage{amssymb}
\usepackage{epsf}

\newcommand{\la}{\langle}
\newcommand{\ra}{\rangle}

\newcommand{\Tr}{\mbox{\bf t}}
\newcommand{\Fa}{\mbox{\bf f}}
\newcommand{\Un}{\mbox{\bf u}}
\newcommand{\Inc}{\mbox{\bf i}}

\newcommand{\HH}[1]{{\cal H}_{#1}}
\newcommand{\HHH}[1]{{\cal H}^4_{#1}}
\newcommand{\hh}[1]{{\cal H}^2_{#1}}
\newcommand{\HHD}[1]{{\cal H}^{dl}_{#1}}

\newcommand{\n}{\mbox{\bf not}}

\newcommand{\vph}{\varphi}

\newcommand{\All}{{\cal A}}

\newcommand{\lar}{\leftarrow}

\newcommand{\WF}{\mathit{WF}}

\newtheorem{theorem}{Theorem}[section]

\title{Uniform semantic treatment of default and autoepistemic logics} 
 
\author{ {\bf Marc Denecker} \\  
Department of Computer Science\\
K.U.Leeuven\\
Celestijnenlaan 200A, B-3001 Heverlee\\
Belgium
\And 
{\bf Victor W. Marek}  \\ 
Computer Science Department\\ 
University of Kentucky\\              
Lexington, KY 40506-0046\\
USA
\And 
{\bf Miros\l aw Truszczy\'nski}   \\ 
Computer Science Department\\
University of Kentucky\\ 
Lexington, KY 40506-0046\\
USA
} 
 
\begin{document} 
 
\maketitle

\begin{abstract} 
We revisit the issue of connections between two leading
formalisms in nonmonotonic reasoning: autoepistemic logic
and default logic. For each logic we develop a comprehensive
semantic framework based on the notion of a {\em belief
pair}. The set of all belief pairs together with the so
called {\em knowledge ordering} forms a complete lattice.
For each logic, we introduce several semantics by means
of fixpoints of operators on the lattice of belief
pairs. Our results elucidate an underlying isomorphism of
the respective semantic constructions. In particular, we
show that the interpretation of defaults as modal formulas
proposed by Konolige allows us to represent all semantics
for default logic in terms of the corresponding semantics
for autoepistemic logic. Thus, our results conclusively
establish that default logic can indeed be viewed as a
fragment of autoepistemic logic. However, as we also
demonstrate, the semantics of Moore and Reiter are given by
{\em different} operators and occupy {\em different} locations
in their corresponding families of semantics! This result
explains the source of the longstanding difficulty to formally
relate these two semantics. In the paper, we also discuss
approximating skeptical reasoning with autoepistemic and
default logics and establish constructive principles behind
such approximations.
\end{abstract} 
 
\section{INTRODUCTION}
\label{intro}

Due to their applications in knowledge representation and, more
specifically, in commonsense reasoning, abduction, diagnosis, belief
revision, planning and reasoning about action, default and autoepistemic
logics are among the most extensively studied nonmonotonic
formalisms\footnote{For a detailed discussion of these two formalisms
and of their applications, and for additional references, the reader
is referred to \cite{mt93}.}. Still, after almost two decades of  
research in the area several key questions remain open.

The first of them is the question of the relationship between 
default and autoepistemic logics. Default logic was introduced by 
Reiter \cite{re80} to formalize reasoning about defaults, that is, 
statements that describe what {\em normally} is the case, in the 
absence of contradicting information. Autoepistemic logic was 
proposed by Moore \cite{mo84} to describe the belief states of 
rational agents reflecting upon their own beliefs and disbeliefs. 
Although the motivation and syntax of both logics are different, 
it has been clear for a long time that they are closely related.
However, despite much work \cite{ko87,mt89a,tru90b,mt93,got95} no
truly satisfactory account of the relationship was found. Konolige
\cite{ko87} related default logic to a version of autoepistemic
logic based on the notion of a {\em strongly grounded expansion} --- a
concept that depends on a syntactic representation of a theory.
Marek and Truszczynski related default logic to two modal nonmonotonic 
logics related but different from autoepistemic logic: the
nonmonotonic modal logic {\bf N} \cite{mt89a} and nonmonotonic modal
logic {\bf S4F} \cite{tru90b}. Finally, Gottlob \cite{got95} found a
relationship between default and autoepistemic logics but the
translation he used was not modular. In fact, he proved that a modular
translation of default logic into autoepistemic logic does not exist. 
These results seem to point to some misalignment between extensions of 
default theories and expansions of modal theories. Our results in 
this paper finally clarify the picture. 

Another problem is related to the fixpoint definitions of extensions
(in the case of default logic) and expansions (in the case of
autoepistemic logic). They provide no insights into constructive
processes agents might use to build their belief sets based on default
or modal theories describing base facts. Finally, there is a problem of
high computational complexity of reasoning with extensions and
expansions. The problems to decide the existence of an extension
(expansion) is $\Sigma_P^2$-complete, the problem to compute the
intersection of extensions (expansions) --- it is needed for skeptical
reasoning --- is $\Pi_P^2$-hard.

In this paper we develop a unifying semantic treatment of default and
autoepistemic logics and use it to address the three issues discussed
above. For each logic we define a family of 2-, 3- and 4-valued semantics
and show that they include all major semantics for default and 
autoepistemic logics. Within our framework we define semantics 
that generalize Kripke-Kleene and 
well-founded semantics for logic programs. These semantics allow us to 
approximate skeptical default and autoepistemic reasoning and they can 
be computed faster than extensions and expansions (assuming that the 
polynomial hierarchy does not collapse). Most importantly, we show that 
the unified semantic picture of default and autoepistemic logics
described here allows us to pinpoint the exact nature of how they
are related.

Our approach is motivated by the algebraic approach proposed 
by Fitting in his analysis of semantics for logic programs with negation
\cite{fi99}, and extends our earlier work on 3-valued semantics 
for autoepistemic logics \cite{dmt98}. It relies on the concept of a
{\em belief pair}, a pair $(P,S)$, where $P$ and $S$ are sets 
of 2-valued interpretations. The concept of a belief pair generalizes
the 
notion of a {\em possible-world structure}, that is, a set of 2-valued 
interpretations, often used to define the 
semantics of the modal logic S5 and also used by Moore and Levesque 
\cite{mo84,le90} in their work on autoepistemic logic. Specifically,
each possible-world structure $Q$ can be identified with a belief pair
$(Q,Q)$. Belief pairs allow us to approximate
the state of beliefs of an agent whose beliefs are represented by 
a possible-world structure.

Given a possible-world structure $Q$ or a belief pair $B$, it is often 
possible to describe a way in which $Q$ (or $B$) could be 
revised to more accurately reflect the agent's beliefs. 
Such a revision procedure can be formally described by an operator.
Fixpoints of such operators are often used to specify semantics of
nonmonotonic formalisms as they represent those belief states of the 
agent that cannot be revised away. Sometimes fixpoints that satisfy some 
minimality conditions are additionally distinguished.

These intuitions underlie the original definition of an {\em expansion}
of a modal theory $T$ as a fixpoint of a certain operator $D_T$ on the set 
of all possible-world structures \cite{mo84}. In \cite{dmt98}, it is shown 
that the operator 
$D_T$ can be extended to the set of belief pairs. The resulting 
operator, ${\cal D}_T$, yields a multi-valued generalization
of expansions and a semantics that approximates 
skeptical autoepistemic reasoning. We refer to it as the Kripke-Kleene 
semantics as it generalizes Kripke-Kleene semantics
for logic programs. In this paper, we derive from the operator 
${\cal D}_T$ two new operators and show that their fixpoints give rise 
to semantics for autoepistemic theories under which circular dependence 
of beliefs upon themselves, present in autoepistemic logic of Moore, 
is eliminated. One of these semantics is shown to correspond to 
Reiter's semantics of default logic. The other one can be viewed as 
a generalization of  the well-founded semantics. 

While possible-world semantics were used in the study of autoepistemic 
logic, they had only a marginal effect on the development of default 
logic. In this paper, we show that possible-world semantics approach 
can be extended to the case of default logic. Namely, in a close 
analogy with autoepistemic logic, for every default theory $\Delta$, 
we introduce an operator ${\cal E}_\Delta$ defined on the set of belief 
pairs. We show that the operator ${\cal E}_\Delta$ gives rise to three 
other operators and that the fixpoints of these operators yield several
semantics for default theories. Among these semantics are the semantics
of extensions by Reiter, the stationary semantics \cite{pp94}, the 
well-founded semantics for default logic \cite{bs91} (it approximates
skeptical reasoning under extensions), the semantics of weak extensions 
\cite{mt89a} and the Kripke-Kleene semantics (it approximates skeptical
reasoning under weak extensions).

Our results settle the issue of the relationship between autoepistemic 
and default logics. We show that the operators ${\cal E}_\Delta$ and 
${\cal D}_T$ are closely related if a default theory $\Delta$ is 
interpreted as a modal theory $T$ given by the translation proposed 
by Konolige \cite{ko87}. Our results show that under the Konolige's 
translation, the {\em families} of semantics for default and autoepistemic 
logics are isomorphic and default logic can be viewed, as has long been 
expected, as a fragment of autoepistemic logic. But this correspondence 
does not relate extensions and expansions! The semantics corresponding
to these concepts occupy different locations in their respective
families of semantics and have different properties. 

We also point out that the Kripke-Kleene and well-founded 
semantics studied in the paper have better computational properties than 
the semantics of expansions and extensions. We conclude with comments 
about further generalizations and open problems.

\section{PRELIMINARIES}
\label{prelim}

The formal language for the semantic study developed in this paper
is that of lattices, operators and fixpoints. The key result is that 
by Tarski and Knaster \cite{ta55} stating that a monotone operator on a 
complete lattice has a least fixpoint.

By $\mathit{At}$ we denote the set of atoms of the propositional 
language under consideration. The set of all 2-valued interpretations of 
$\mathit{At}$ will be denoted by ${\cal A}$. Any set $Q \subseteq
\All$ is called a {\em possible-world} structure and can be 
viewed as a universal Kripke model \cite{che80}. Possible-world 
structures are a basic tool in semantic studies of modal logics. They 
were also used in the context of autoepistemic logic \cite{mo84,le90,mt93}. 

A collection of all possible-world structures will be denoted by 
${\cal W}$. This set can be ordered by the reverse set 
inclusion: for $Q_1,Q_2 \in{\cal W}$, $Q_1\sqsubseteq Q_2$ if 
$Q_2 \subseteq Q_1$ (the smaller the set of possible worlds, the bigger
the theory it determines). Clearly, $\la {\cal W}, \sqsubseteq \ra$ is a 
complete lattice. To study formalisms based on the modal language, we 
define the {\em truth function} $\HH{Q,I}$ inductively as follows
($Q$ is a possible-world structure, $I\in \All$ is an interpretation):
\begin{enumerate}
\item $\HH{Q,I}(p) = I(p)$, if $p$ is an atom
\item Boolean connectives are handled in the standard Tarskian way
\item $\HH{Q,I}(K\vph) = \Tr$, if for every interpretation $J\in Q$, 
$\HH{Q,J}(\vph)=\Tr$, and $\HH{Q,I}(K\vph) = \Fa$, otherwise.
\end{enumerate}
The value of a modal atom $K\vph$ given by $\HH{Q,I}$
does not depend on $I$. Thus, it is determined only by the
possible-world structure in question. It is clear that
the {\em meta-knowledge} specified by a possible-world structure $Q$
is complete: all modal atoms $K\vph$ are either true or false with
respect to $Q$. 

For every modal theory $T$, Moore \cite{mo84} defined an operator $D_T$ 
on ${\cal W}$ by:
\[
D_T(Q) = \{I: \HH{Q,I}(\vph) = \Tr,\ \mbox{for every $\vph\in T$}\}.
\]
Moore called the theory of a fixpoint of $D_T$ an {\em expansion}\footnote{In
the paper, we will frequently use the same term to denote the fixpoint 
of an operator and its theory. So, fixpoints of $D_T$ will be referred
to as expansions, as well.}.

In \cite{dmt98}, Moore's approach was extended to the 3-valued case, 
in which a possibility of incomplete meta-knowledge is admitted. A key 
concept is that of a {\em belief pair}, that is, a pair $(P,S)$ of
possible-world structures. In a belief pair $(P,S)$, $P$ can be viewed
as a representation of a conservative (pessimistic) view on what is
believed while $S$ can be regarded as a representation of a liberal
(gullible) view. If $S\subseteq P$, formulas believed in according
to the conservative view captured by $P$, are believed in according to  
the liberal view represented by $S$. Consequently, a belief pair 
$(P,S)$ such that $S\subseteq P$ is called {\em consistent}. However, 
the ways in which the agent establishes the estimates $P$ and $S$
may be independent of each other and, thus, we allow belief pairs $(P,S)$ 
such that $S$ is not a subset of $P$. We
refer to them as {\em inconsistent}\footnote{In \cite{dmt98} only
consistent belief pairs were considered.}. In applications we are mostly
interested in consistent belief pairs. Constructive techniques for
building belief pairs given a base theory, which are described in the
paper, result in consistent belief pairs. However, admitting inconsistent 
belief pairs completes the picture, leads to simple intuitions behind
mathematical arguments and results in more elegant algebraic structures. 

With a belief pair $(P,S)$ and a 2-valued interpretation $I$ we 
associate a 2-valued truth function $\hh{(P,S),I}$ defined on a modal
language. The idea is to define $\hh{(P,S),I}(\vph)$ so that it provides a 
conservative estimate to the truth value of $\vph$ with respect to a 
belief pair $(P,S)$. Since $P$ represents a conservative point of view and
$S$ a liberal one, to get a conservative estimate we use $P$ to 
evaluate positive occurrences of
modal atoms and $S$ to evaluate negative ones. The definition is 
inductive:
\begin{enumerate}
\item $\hh{(P,S),I}(p)=I(p)$, for every atom $p$
\item Conjunction and disjunction are treated in the standard Tarskian way
\item $\hh{(P,S),I}(\neg \vph) = \neg{\hh{(S,P),I}(\vph)}$
\item $\hh{(P,S),I}(K\vph) = \Tr$ if $\hh{(P,S),J}(\vph)=\Tr$ for
all $J \in P$. $\hh{(P,S),I}(K\vph) = \Fa$, otherwise.
\end{enumerate}

We stress that when evaluating the negation of a formula the 
roles of $P$ and $S$ are switched, which ensures that modal literals 
appearing positively in a formula are evaluated with respect to $P$ while 
those appearing negatively are evaluated with respect to $S$.

Clearly, to construct a liberal estimate for the truth value of 
$\vph$ with respect to a belief pair $(P,S)$ we can proceed similarly 
and use
$S$ ($P$) to evaluate modal literals appearing positively (negatively)
in $\vph$. It is easy to see, however, that the resulting truth function 
can be expressed as $\hh{(S,P),I}$ (we reverse the roles of $P$ and $S$). 

Conservative and liberal estimates of truth values of formulas can be
combined into a single {\em 4-valued} estimate. We say that the logical
value of a formula $\vph$ is true, $\Tr_4$ (false, $\Fa_4$), if both 
conservative and liberal 
estimates of its truth value are equal to $\Tr$ ($\Fa$). We say that the
logical value $\vph$ is {\em unknown}, $\Un$, if the conservative estimate 
is $\Fa$ and the liberal estimate is $\Tr$. Finally, the logical value of
$\vph$ is {\em inconsistent}, $\Inc$, if the conservative estimate 
is $\Tr$ and the liberal estimate is $\Fa$. Thus, the estimates given by 
$\hh{(P,S),I}(\vph)$ and $\hh{(S,P),I}(\vph)$ yield a 4-valued truth function
\[
\HHH{(P,S),I}(\vph) = (\hh{(P,S),I}(\vph), \hh{(S,P),I}(\vph)),
\]
where
$\Tr_4 = (\Tr,\Tr)$,  $\Fa_4 = (\Fa,\Fa)$, $\Un=(\Fa,\Tr)$ and
$\Inc=(\Tr,\Fa)$.

We define the {\em meta-knowledge} of a belief pair $(P,S)$ as the set
of all formulas $\vph$ such that both conservative and liberal estimates
of the truth value of the epistemic atom $K\vph$ coincide (are both true
or are both false). Clearly, the meta-knowledge of a belief pair $(P,S)$ 
need not be complete. For some modal atoms $K\vph$ the two estimates
may disagree ($K\vph$ may be assigned value $\Un$ or $\Inc$). However,
it is easy to see, that for a complete belief pair $(P,P)$,
$\HHH{(P,P),I} =\HH{P,I}$, for every interpretation $I\in \All$. In
other words, belief pairs and the truth function $\HHH{(P,S),I}$
generalize possible-world structures and the truth function $\HH{P,I}$.
In addition, for a consistent belief pair $(P,S)$, $\HHH{(P,S),I}$ never
assigns value $\Inc$ and coincides with the 3-valued truth function
defined in \cite{dmt98}.
 
The set of all belief pairs is denoted by $\cal B$. It can be ordered by 
the {\em knowledge ordering} $\preceq_{kn}$: $(P_1, S_1) \preceq_{kn} 
(P_2, S_2)$ if $S_1 \subseteq S_2$, and $P_2 \subseteq P_1$. The
set $\cal B$ with the ordering $\preceq_{kn}$ forms a complete
lattice and, consequently, a $\preceq_{kn}$-monotone operator on ${\cal B}$
is guaranteed to have a least fixpoint by the result of Tarski and Knaster
\cite{ta55}.
The ordering $\preceq_{kn}$ coincides with the ordering of increasing
meta-knowledge (decreasing meta-ignorance): $(P_1, S_1)
\preceq_{kn} (P_2, S_2)$ if and only if the set of modal atoms with 
the same conservative and liberal estimates with respect to $(P_1,S_1)$ 
is contained in the set of modal atoms for which conservative and
liberal estimates with respect to $(P_2,S_2)$ are the same. 

Let $(P,S)$ be a belief pair. Following \cite{dmt98}, we set
\[
{\cal D}_T(P,S) = ({\cal D}^l_T(P,S),{\cal D}^u_T(P,S)),
\]
where 
\[
{\cal D}^l_T(P,S)= \{I\colon \hh{(S,P),I}(T)= \Tr\}
\]
and 
\[
{\cal D}^u_T(P,S)=\{I\colon \hh{(P,S),I}(T) = \Tr\}.
\]
Fixpoints of the operator ${\cal D}_T$ will be called {\em partial 
expansions}.  

Speaking intuitively, the operator ${\cal D}_T$ describes how an agent 
might revise a belief pair $(P,S)$. The possible-world structure $P$
is replaced by the structure $P'$ consisting of those interpretations 
$I$ for which all formulas from $T$ are true according to the liberal 
estimates of truth values (given $(P,S)$). A liberal criterion for 
selecting possible worlds (interpretations) to $P'$ results in a 
possible-world structure capturing a conservative point of view.
By duality between conservative and liberal approaches, $S$ can be 
replaced by $S'$ consisting of all interpretations $I$ such that all 
formulas from $T$ are true according to the conservative estimates of 
truth values (given $(P,S)$). 

Let us recall that $D_T$ stands for the operator introduced by Moore. 
The operator ${\cal D}_T$ allows us to reconstruct the semantic approach 
to autoepistemic logic proposed by Moore.

\begin{theorem}\label{SymApp}
Let $T$ be a modal theory. Then, for every possible-world structure $P$, 
${\cal D}_T(P,P) = (D_T(P),D_T(P))$. Consequently, a belief pair $(P,P)$ 
is a fixpoint of ${\cal D}_T$ if and only if $P$ is a fixpoint of $D_T$.
\end{theorem}

Theorem \ref{SymApp} implies that there is a natural one-to-one correspondence
between the complete partial expansions of $T$ and the expansions of $T$.
Thus, partial expansions (consistent partial expansions) can be viewed as 
4-valued (3-valued) generalizations of Moore's expansions.

The key property of the operator ${\cal D}_T$ is its
$\preceq_{kn}$-monotonicity. It follows that ${\cal D}_T$ has a unique
$\preceq_{kn}$-least fixpoint. We denote it by $KK(T)$ and refer to 
it as the {\em
Kripke-Kleene fixpoint} (or semantics) for $T$\footnote{There is
a close analogy between the least fixpoint of the operator ${\cal D}_T$
and the Kripke-Kleene semantics for logic programs.}. Kripke-Kleene
fixpoint has a clear constructive flavor (it can be obtained by iterating
the operator ${\cal D}_T$, starting at the least informative belief
pair, $(\All , \emptyset )$). It approximates all partial expansions and, 
in particular, approximates the skeptical reasoning with expansions. 

\begin{theorem}
Let $T$ be a modal theory.
\begin{enumerate}
\item The fixpoint $KK(T)$ is consistent.
\item  For every partial expansion $B$ of $T$, $KK(T)\preceq_{kn} B$.
\item If $K\vph$ is true with respect to $KK(T)$ then $\vph$
belongs to every expansion of $T$. If $K\vph$ is false with respect to
$KK(T)$ then $\vph$ belongs to no expansion of $T$.
\end{enumerate}
\end{theorem}

Deciding the truth value of a modal atom $K\vph$ with respect to the
Kripke-Kleene fixpoint is in the class $\Delta_P^2$ \cite{dmt98}. Thus, 
unless the polynomial hierarchy collapses, it is a simpler problem
than the problem of computing expansions or their intersection.

We can also use the Kripke-Kleene semantics as a test for the uniqueness of
an expansion. Namely, we can first compute $KK(T)$ and check if $KK(T)$
is complete. If it is, $T$ has a unique expansion. This method is
computationally better (again, if the polynomial hierarchy does not
collapse) than the straightforward one which computes all expansions
of $T$. However, it is incomplete --- there are theories $T$ with a 
unique expansion and such that $KK(T)$ is not complete.

\section{AUTOEPISTEMIC LOGIC}\label{ael.sec}

The operator ${\cal D}_T$ allows us to
define two additional operators: the operator ${D}_T^{st}$ defined on 
the lattice ${\cal W}$, and the operator ${\cal D}_T^{st}$ defined on the 
lattice ${\cal B}$. They give rise to new semantics for autoepistemic 
logic that are closely related to the semantics of extensions for
default logic.   
One of them is a perfect match to Reiter's semantics of extensions 
for default logic, an object long sought after in the autoepistemic logic.

Let us recall that the operator ${\cal D}_T$ associates with each belief
pair $(P,S)$ its revised variant $(P',S') = {\cal D}_T(P,S)$. The way in
which $P'$ is obtained is described by the operator ${\cal D}_T^l$.
Namely, $P' = {\cal D}_T^l(P,S)$. If we fix $S$, this operator becomes 
a monotone operator on ${\cal W}$ (it follows from the fact that 
${\cal D}_T$ is $\preceq_{kn}$-monotone). Hence, its least fixpoint 
can be viewed as the preferred revision of $P$, given a fixed $S$. Let
us define, then,
\[
D_T^{st}(S) = \mathit{lfp}({\cal D}_T^l(\cdot, S)).
\]
Similarly, we can argue that $\mathit{lfp}({\cal D}_T^u(P,\cdot))$ can
be regarded as a preferred revision of $S$, given $P$. It turns out that
$\mathit{lfp}({\cal D}_T^u(P,\cdot)) = D^{st}_T(P)$. Thus, we define an
operator ${\cal D}_T^{st}$ on belief pairs as follows:
\[
{\cal D}_T^{st}(P,S) = (D_T^{st}(S),D_T^{st}(P)).
\]

Clearly, $D^{st}_T$ is an operator on ${\cal W}$. The fixpoints of the 
operator $D_T^{st}$ (and also their theories) will be referred to as 
{\em extensions}. The choice of the term is not arbitrary. We show in 
Section \ref{connect} that extensions of modal theories can be regarded 
as generalizations of extensions of default theories.
We have the following property relating fixpoints of the operators
$D_T^{st}$ and ${\cal D}_T^{st}$.

\begin{theorem}
For every modal theory $T$, a possible-world structure $P$ is a fixpoint
of $D_T^{st}$ if and only if a belief pair $(P,P)$ is a fixpoint of
${\cal D}_T^{st}$.
\end{theorem}

It follows that the semantics of fixpoints of ${\cal D}^{st}_T$ can be
viewed as a 4-valued version of the semantics of extensions. Similarly,
consistent fixpoints can be thought of as a 3-valued generalizations of
extensions. Consequently, we refer to the fixpoints of the operator
${\cal D}_T^{st}$ as {\em partial extensions}.

The circular dependence allowing the agent to accept $p$ to the belief set 
just on the basis of this agent believing in $p$, allowed under the
semantics of expansions, is eliminated in the case of
extensions. For instance, the theory $\{Kp\Rightarrow p\}$ has 
two expansions. One of them is determined by the possible-world structure 
consisting of all interpretations, the other one --- by the possible-world 
structure consisting of all interpretations in which $p$ is true. It is 
this second expansion that suffers from circular-argument problem: the 
belief in $p$ is the only justification for having $p$ in this expansion. 
In the same time, the theory $\{Kp\Rightarrow p\}$ has exactly {\em one}
extension, the one given by the possible-world structure consisting of 
all interpretations. The atom $p$ is not true in it and, hence, circular
arguments are not used in the construction of this expansion.

The operator $D_T^{st}$ is antimonotone. It follows that the
operator ${\cal D}_T^{st}(P,S)$ is $\preceq_{kn}$-monotone. Thus,
by the result of Tarski and Knaster, it has the least fixpoint. We will 
denote this fixpoint by $\WF(T)$ and refer to it as the {\em well-founded 
fixpoint} (or semantics) of $T$. Our choice of the term is again 
not accidental. This semantics is closely related to the well-founded 
semantics of default logic \cite{bs91} and logic programming \cite{vrs91}. 
We have the following result
indicating that well-founded semantics can be used to approximate
all partial extensions and, in particular, all extensions of a modal
theory. It also shows that the well-founded semantics provides
a sufficient condition for the uniqueness of an extension.

\begin{theorem}
Let $T$ be a modal theory and let $\WF(T)=(P,S)$. Then:
\begin{enumerate}
\item The fixpoint $\WF(T)$ is consistent.
\item For every partial extension $B$, $\WF(T) \preceq_{kn} B$.
\item If $\HHH{\WF(T),I}(K\vph)=\Tr$, then $\vph$ belongs to every 
extension of $T$. Similarly, if $\HHH{\WF(T),I}(K\vph)=\Fa$, then $\vph$ 
does not belong to any extension.
\item If $\WF(T)$ is complete (that is, $P=S$) then $T$ has a unique
extension corresponding to the possible-world structure $P$.
\end{enumerate}
\end{theorem}

Based on the approach developed in \cite{dmt98} for computing the
Kripke-Kleene semantics, we can establish computational properties
of the well-founded semantics. We have the following result.

\begin{theorem}
The problem of computing the well-founded semantics is the class
$\Delta_P^2$.
\end{theorem}

Thus, assuming that the polynomial hierarchy does not collapse, 
computing the well-founded semantics of a theory $T$ is easier than 
computing the intersection of extensions of $T$ (that is, the set 
of skeptical consequences of $T$). Since the well-founded semantics 
of a modal theory $T$ approximates all extensions, it can be used to 
speed up the computation of their intersection. 

The next result connects expansions and the Kripke-Kleene semantics
with extensions and the well-founded semantics. It shows that the
well-founded semantics is stronger than the Kripke-Kleene semantics and
that (partial) extensions of $T$ are (partial) expansions of $T$
satisfying some minimality condition. 

\begin{theorem}
Let $T$ be a modal theory. Then:
\begin{enumerate}
\item $KK(T) \preceq_{kn} \WF(T)$.
\item Every extension of $T$ is a $\sqsubseteq$-minimal expansion
of $T$.
\item Every partial extension $(P,S)$ of $T$ is a minimal partial 
expansion of $T$ in the following sense: for every partial expansion 
$(P',S')$, if $P'\sqsubseteq P$ and $S'\sqsubseteq S$, then $P=P'$ 
and $S=S'$.
\end{enumerate}
\end{theorem} 

We conclude this section with a schematic illustration of the panorama
of semantics for autoepistemic logic. The central position is occupied 
by the operator ${\cal D}_T$. Its fixpoints yield the semantics of 
partial expansions and its least fixpoint yields the Kripke-Kleene
semantics. Restriction of the operator ${\cal D}_T$ to complete belief
pairs leads to the operator $D_T$, originally introduced by Moore, and
results in the semantics of expansions. The operator ${\cal D}_T$
also gives rise to the operators $D_T^{st}$ and ${\cal D}_T^{st}$ that
yield new semantics for autoepistemic logic: the semantics of
extensions, the semantics of partial extensions and the well-founded 
semantics.

\begin{figure}[t]
\begin{center}
\begin{picture}(60,150)(-30,-40)
\put(-100,-5){\scriptsize AE logic of Moore}
\put(-85,-15){\scriptsize expansions}
\put(-50,10){\vector(1,1){10}}
\put(-100,95){\scriptsize partial expansions}
\put(-110,85){\scriptsize Kripke-Kleene fixpoint}
\put(-20,80){\vector(1,-1){10}}
\put(0,60){\makebox(0,0){${\cal D}_T$}}
\put(-30,30){\makebox(0,0){${D}_T$}}
\put(20,-30){\scriptsize partial extensions}
\put(10,-40){\scriptsize well-founded semantics}
\put(50,50){\vector(-1,-1){10}}
\put(55,58){\scriptsize extensions}
\put(20,-20){\vector(-1,1){10}}
\put(30,30){\makebox(0,0){$D^{st}_T$}}
\put(0,0){\makebox(0,0){${\cal D}^{st}_T$}}
\put(-7,53){\vector(-1,-1){17}}
\put(10,10){\vector(1,1){13}}
\put(20,20){\vector(-1,-1){13}}
\put(7,53){\vector(1,-1){17}}
\end{picture}
\end{center}
\caption{Operators associated with autoepistemic logic}
\label{fig1}
\end{figure}
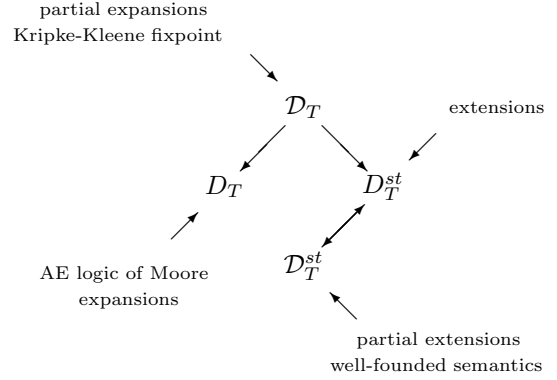

\section{DEFAULT LOGIC}

While possible-world semantics played a prominent role in the study of
autoepistemic logics \cite{mo84,le90,dmt98} they have not, up to now,
had a similar impact on default logic. In this section we will
introduce a comprehensive semantic treatment of default logic
in terms of possible-world structures and belief pairs. Our approach
will follow closely that used in the preceding sections.
 
We observed earlier that autoepistemic logic can be viewed as the logic 
of the operator ${\cal D}_T$. Its fixpoints, and fixpoints of the 
operators that can be derived from ${\cal D}_T$, determine all major
semantics for autoepistemic logic. We will now develop a similar
treatment of default logic. 

As before, we start with a 2-valued truth function that gives a
conservative estimate of the logical value of a formula or a default 
with respect to a belief pair $(P,S)$ and an interpretation $I$. For 
a propositional formula $\vph$, we define $\HHD{(P,S),I}(\vph) = I(\vph)$. 
For a default $d = \frac{\alpha\colon \beta_1,\ldots,\beta_k}{\gamma}$, 
we set $\HHD{(P,S),I}(d)=\Tr$ if at least one of the following
conditions holds:
\begin{enumerate}
\item there is $J\in S$ such that $J(\alpha)=\Fa$
\item there is $i$, $1\leq i\leq k$ such that for every $J\in P$,
$J(\beta_i)=\Fa$
\item $I(\gamma)=\Tr$.
\end{enumerate}
We set $\HHD{(P,S),I}(d)=\Fa$, otherwise. Clearly, the definition of
$\HHD{(P,S),I}(d)$ agrees with the intuitive reading of a default $d$:
it is true, according to a conservative point of view, if its
prerequisite is false (even with respect to a liberal view captured 
by $S$) or if at least one of its justifications is definitely impossible 
(it is false according to a conservative point of view captured by $P$) or
if its consequent is true (in $I$). As before, we can also argue that 
$\HHD{(S,P),I}(d)$ provides a liberal estimate for a truth 
value of $d$ with respect to $(P,S)$ (the roles of $P$ and $S$ are reversed). 

Let $\Delta = (D, W)$ be a default theory.
We use the truth function $\HHD{(P,S),I}$ to define an operator 
${\cal E}_{\Delta}$ on the lattice $\cal B$ of belief pairs:
\[
{\cal E}_\Delta (P,S) = ({\cal E}^{l}_\Delta(P,S),{\cal
E}^{u}_\Delta(P,S)),
\]
where 
\[
{\cal E}^{l}_\Delta(P,S) = \{I\colon \HHD{(S,P),I}(\Delta)=
\Tr\}
\]
and 
\[
{\cal E}^{u}_\Delta(P,S)=\{I\colon \HHD{(P,S),I}(\Delta)=
\Tr\}.
\]
This definition can be justified similarly as that of the
operator ${\cal D}_T$ in Section \ref{ael.sec}. 

We will now define a 2-valued version of the operator ${\cal E}_\Delta$.
To this end, we use the following result.

\begin{theorem}\label{ttt-14}
Let $\Delta$ be a default theory. If $B\in{\cal B}$ is complete then 
${\cal E}_\Delta(B)$ is also complete.
\end{theorem} 

Let $Q$ be a possible-world structure. We define
\[
E_\Delta(Q)=Q',
\] 
where $Q'$ is a possible-world structure such that   
${\cal E}_\Delta(Q,Q) = (Q',Q')$ (its existence is guaranteed by 
Theorem \ref{ttt-14}).  

To the best of our knowledge, the operator $E_\Delta$ has not appeared 
explicitly in the literature before. Its fixpoints, however, did. In 
\cite{mt89a}, the concept of a {\em weak extension} of a default theory 
was introduced and studied (the approach used there was proof-theoretic). 
It turns out that fixpoints of the operator $E_\Delta$ correspond precisely 
to weak extensions of $\Delta$. Thus, the semantics given by the operator 
$E_\Delta$ is precisely the semantics of weak extensions.

\begin{theorem}\label{ttt}
Let $\Delta$ be a default theory. Then:
\begin{enumerate}
\item A propositional theory $T$ is a weak extension of a default theory 
$\Delta$ according to {\rm \cite{mt89a}} if and only if 
$T = \{\vph\colon I(\vph) =\Tr, \ \mbox{for every $I\in Q$}\}$
for a fixpoint $Q$ of $E_{\Delta}$.
\item A possible-world structure $Q$ is a fixpoint of $E_\Delta$ if
and only if a belief pair $(Q,Q)$ is a fixpoint of ${\cal E}_\Delta$. 
\end{enumerate}
\end{theorem}

In view of Theorem \ref{ttt}(1), we call fixpoints of the operator
$E_{\Delta}$ {\em weak extensions}. Theorem \ref{ttt}(2) implies that 
complete fixpoints of the operator ${\cal E}_\Delta$ are in one-to-one 
correspondence with the fixpoints of the operator $E_\Delta$. Thus,
we call fixpoints of the operator ${\cal E}_\Delta$ --- {\em partial weak 
extensions} (they can be regarded as a 4-valued generalization
of weak extensions, consistent fixpoints can be regarded as 3-valued
generalizations).

The key property of the operator ${\cal E}_{\Delta}$ is its
$\preceq_{kn}$-monotonicity. Thus, ${\cal E}_\Delta$ has a least fixpoint.
We call it the {\em Kripke-Kleene fixpoint} and denote it by $KK(\Delta)$. 
We refer to the corresponding semantics as the {\em Kripke-Kleene
semantics} for $\Delta$.

The Kripke-Kleene semantics can be obtained by iterating the operator 
${\cal E}_\Delta$ starting with the least informative belief pair 
$(\All, \emptyset)$. Thus, it has a constructive flavor. Second, it 
approximates the skeptical reasoning with weak extensions and provides 
a test for uniqueness of a weak extension.

\begin{theorem}\label{ttt-2}  
Let $\Delta$ be a default theory and let $KK(\Delta)=(P,S)$. 
\begin{enumerate}
\item The fixpoint $KK(\Delta)$ is consistent, that is, $S\subseteq
P$.
\item If $I(\vph)=\Tr$ for every $I\in P$, then $\vph$ belongs to every 
weak extension. If $J(\vph)=\Fa$ for some $J\in S$, $\vph$ does not belong to
any weak extension.
\item If $P=S$ (that is, if $KK(\Delta)$ is complete) then $\Delta$ has 
a unique weak extension corresponding to the possible-world structure $P$.
\end{enumerate}
\end{theorem}

The Kripke-Kleene semantics is computationally attractive. 
Deciding whether a formula is in the intersection of 
the weak extensions of a default theory $\Delta$ is $\Pi_P^2$-complete.
In contrast, by adapting the methods developed in \cite{dmt98} to the 
case of default logic we can show that the problem of computing 
$KK(\Delta)$ (specifically, assuming $KK(\Delta)=(P,S)$, deciding 
whether $I(\vph)=\Tr$ for every $I\in P$, or whether $J(\vph)=\Fa$ for 
some $J\in S$) is in the class $\Delta_P^2$. 

So far we have not yet reconstructed the concept of an extension. In 
order to do so, we will now derive from ${\cal E}_\Delta$ two other
operators related to default logic. Let us consider a belief pair
$(P,S)$. We want to revise it to a belief pair $(P',S')$. We might do it
by fixing $S$ and taking for $P'$ a preferred revision of $P$, and by
fixing $P$ and taking for $S'$ a preferred revision of $S$.

It is easy to see that $\preceq_{kn}$-monotonicity of ${\cal E}_\Delta$ 
implies that the operator ${\cal E}^l_\Delta(\cdot,S)$ is 
$\sqsubseteq$-monotone operator on $\cal W$. Consequently, it has a
least fixpoint. This fixpoint can be taken as the preferred way
to revise $P$ given $S$. Thus, we define
\[
E_\Delta^{st}(S) = \mathit{lfp}(({\cal E}_\Delta)^l(\cdot, S))
\]
As in the case of autoepistemic logic, one can see that
$E_\Delta^{st}$ also specifies the preferred way to revise 
$S$ given $P$, that is $E_\Delta^{st}(P) = \mathit{lfp}({\cal
E}_\Delta^u(P, \cdot))$. Thus, we define the operator on $\cal B$ as
follows:
\[
{\cal E}^{st}_\Delta(P,S) = (E_\Delta^{st}(S), E_\Delta^{st}(P)).
\]

It turns out that the concept of extension as defined by Reiter can be
obtained from the operator $E_\Delta^{st}$. It is known
that Reiter's extensions are theories of fixpoints of the operator
$\Sigma_\Delta$ introduced by Guerreiro and Casanova \cite{gc90}
(see also \cite{li90a,mt93}). One can show that the operator 
$E_\Delta^{st}$ coincides with the operator $\Sigma_\Delta$.
Thus, we have the following result.

\begin{theorem}\label{defa}
A theory $T$ is an extension of a default theory $\Delta$ if and only if
$T = \{\vph\colon I(\vph) =\Tr, \ \mbox{for every $I\in Q$}\}$ for
some fixpoint $Q$ of $E_\Delta^{st}$.
\end{theorem}

In view of Theorem \ref{defa}, we refer to the fixpoints of $E_\Delta^{st}$ 
as {\em extensions}. We have the following result relating fixpoints of the 
operators $E_\Delta^{st}$ and ${\cal E}^{st}_\Delta$.

\begin{theorem}\label{ttt-18}
Let $\Delta$ be a default theory. For every possible-world structure
$P$, $P$ is a fixpoint of $E_\Delta^{st}$ if and only if $(P,P)$ is a 
fixpoint of ${\cal E}^{st}_\Delta$.
\end{theorem}

It follows that the fixpoints of ${\cal E}^{st}_\Delta$ can be regarded
as 4-valued (3-valued, in the case of consistent fixpoints) generalizations 
of an extension of a default theory. We will therefore
call them {\em partial extensions}. It turns out that partial extensions
coincide with stationary extensions defined in \cite{pp94}.

Our next result describes monotonicity properties of the operators
$E_\Delta^{st}$ and ${\cal E}^{st}_\Delta$.

\begin{theorem}\label{ttt-3}
Let $\Delta$ be a default theory. Then, the operator 
$E_\Delta^{st}$ is $\sqsubseteq$-antimonotone and the operator
${\cal E}^{st}_\Delta$ is $\preceq_{kn}$-monotone.
\end{theorem}

Theorem \ref{ttt-3} implies that the operator ${\cal
E}^{st}_\Delta$ has a least fixpoint. We will denote it by $\WF(\Delta)$
and refer to it as the {\em well-founded fixpoint} of $\Delta$. We will 
call the semantics it implies a {\em well-founded semantics} of $\Delta$.
The well-founded semantics of $\Delta$ coincides with the well-founded
semantics of default logic introduced by Baral and Subrahmanian \cite{bs91}.
The well-founded semantics allows us to approximate skeptical reasoning
with extensions and yields a sufficient condition for the uniqueness 
of an extension.

\begin{theorem}\label{opEst4}
Let $\Delta$ be a default theory and let $\WF(\Delta)=(P,S)$. Then:
\begin{enumerate}
\item The fixpoint $\WF(\Delta)$ is consistent.
\item For every partial extensions $B$ of ${\cal E}^{st}_\Delta$,
$\WF(\Delta)\preceq_{kn} B$.
\item If $J(\vph)=\Tr$ for every $I\in P$, then $\vph$ belongs to
every extension. If $J(\vph)=\Fa$ for some $J\in S$, $\vph$ does not
belong to any extension.
\item If $P=S$ (that is, if $\WF(\Delta)$ is complete) then $\Delta$ 
has a unique extension corresponding to the possible-world structure $P$.
\end{enumerate}
\end{theorem}

Well-founded semantics has a constructive flavor. It can be obtained by
iterating the operator ${\cal E}^{st}_\Delta$ over the belief pair
$(\All,\emptyset)$. In addition, by extending the approach described in 
\cite{dmt98}, one can show that the problem of computing the well-founded 
semantics is in the class $\Delta_P^2$. 

Finally, let us note connections between (partial) weak extensions and
(partial) extensions, and between the Kripke-Kleene and well-founded 
semantics for default logic.

\begin{theorem}
Let $\Delta$ be a default theory. Then:
\begin{enumerate}
\item $KK(\Delta) \preceq_{kn} \WF(\Delta)$.
\item Every extension of $\Delta$ is a $\sqsubseteq$-minimal
weak extension of $\Delta$.
\item Every partial extension $(P,S)$ of $\Delta$ is a minimal 
partial weak extension of $\Delta$ in the following sense: for every 
partial weak extension $(P',S')$, if $P'\sqsubseteq P$ and $S'\sqsubseteq S$, 
then $P=P'$ and $S=S'$.
\end{enumerate}
\end{theorem}

\begin{figure}[t]
\begin{center}
\begin{picture}(60,150)(-30,-40)
\put(-100,-5){\scriptsize weak extensions}
\put(-50,10){\vector(1,1){10}}
\put(-100,95){\scriptsize Kripke-Kleene fixpoint}
\put(-20,80){\vector(1,-1){10}}
\put(0,60){\makebox(0,0){${\cal E}_\Delta$}}
\put(-30,30){\makebox(0,0){${E}_\Delta$}}
\put(20,-30){\scriptsize partial extensions}
\put(10,-40){\scriptsize well-founded semantics}
\put(50,50){\vector(-1,-1){10}}
\put(45,58){\scriptsize Reiter's extensions}
\put(20,-20){\vector(-1,1){10}}
\put(30,30){\makebox(0,0){$E^{st}_\Delta$}}
\put(0,0){\makebox(0,0){${\cal E}^{st}_\Delta$}}
\put(-7,53){\vector(-1,-1){17}}
\put(10,10){\vector(1,1){13}}
\put(20,20){\vector(-1,-1){13}}
\put(7,53){\vector(1,-1){17}}
\end{picture}
\end{center}
\caption{Operators associated with default logic}
\label{fig2}
\end{figure}
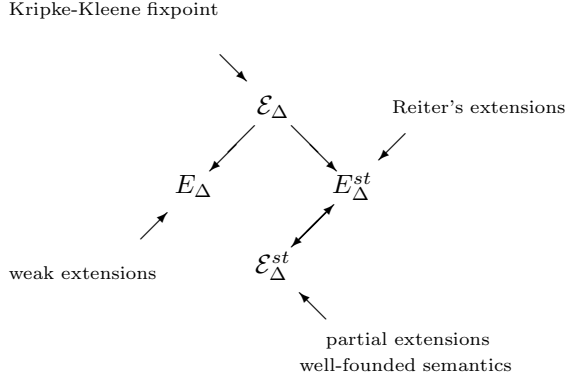

In summary, default logic can be viewed as the logic of the operator
${\cal E}_\Delta$. Its fixpoints define the semantics of partial weak 
extensions. The least fixpoint of ${\cal E}_\Delta$ defines the 
Kripke-Kleene semantics. The operator ${\cal E}_\Delta$ gives rise to 
the operator $E_\Delta$, which yields the semantics of weak extensions. 
Kripke-Kleene semantics provides an approximation for the skeptical 
reasoning under the semantics of weak extensions. The operator 
${\cal E}_\Delta$ also leads to the operator ${\cal E}^{st}_\Delta$. 
Consistent fixpoints of this operator yield stationary extensions. 
Fixpoints of a related operator $E_\Delta^{st}$, defined on the lattice 
${\cal W}$, correspond to extensions by Reiter. The least fixpoint of 
the operator ${\cal E}^{st}_\Delta$ results in the well-founded semantics 
for default logic and approximates the skeptical reasoning under the 
semantics of extensions. The relationships between the operators of
default logic are illustrated in Figure \ref{fig2}.

\section{DEFAULT LOGIC VERSUS AUTOEPISTEMIC LOGIC}\label{connect}

The results of the paper shed new light on the relationship between
default and autoepistemic logics. The nature of this relationship was
the subject of extensive investigations since the time both systems
were introduced in early 80s. Konolige \cite{ko87} proposed to encode
a default $d=\frac{\alpha\colon \beta_1,\ldots,\beta_k}{\gamma}$ by the
modal formula
$m(d) = K\alpha \wedge \neg K\neg \beta_1 \wedge \ldots \neg K\neg
\beta_k \supset \gamma$
and to represent a default theory $\Delta = (D,W)$ by a modal theory
$m(\Delta ) = W \cup \{m(d)\colon d\in D\}$. Despite the fact that the
encoding is intuitive it does not provide a correspondence between
default logic as defined by Reiter and autoepistemic logic as defined by
Moore. Consider a default
theory $\Delta$ with $W=\emptyset$ and $D=\{\frac{p\colon q}{p}\}$, where 
$p$ and $q$ are two different atoms. Then $\Delta$ has exactly one extension,
$Cn(\emptyset)$. Applying the translation of Konolige to $\Delta$ yields
the theory $m(\Delta )= \{Kp\wedge \neg K\neg q \supset p\}$.
The theory $m(\Delta )$ has two expansions. One of them is generated by the
theory $Cn(\emptyset)$ and corresponds to the only extension of $\Delta$.
The other expansion is generated by the theory $Cn(\{p\})$. Thus, the
Konolige's translation does not give a one-to-one correspondence between
extensions of default theories and expansions of their modal encodings.

This mismatch can be explained within the semantic framework
introduced in the paper. Konolige's translation does not establish
correspondence between extensions and expansions because they are
associated with different operators. Expansions are associated with
fixpoints of the operator $D_T$. Its counterpart on the side of
default logic is the operator $E_{\Delta}$. Fixpoints of this operator
are not extensions but weak extensions of $\Delta$. Extensions turn out
to be associated with the operator $E^{st}_\Delta$. Its counterpart on
the side of autoepistemic logic is the operator $D^{st}_T$, introduced
in Section \ref{ael.sec}. This operator, to the best of our knowledge,
has not appeared in the literature and properties of its fixpoints and
the relationship to McDermott-Doyle style logics \cite{mcdd80,mcd82,mt93} 
are not known.

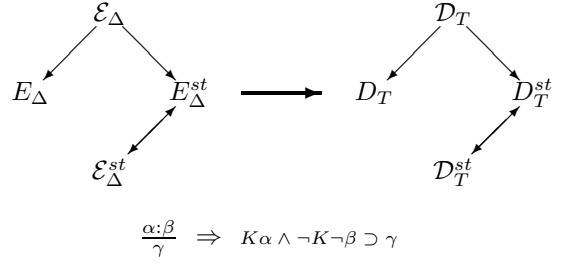
\begin{figure}[t]
{
\begin{center}
\begin{picture}(60,105)(0,-30)
\put(-40,60){\makebox(0,0){${\cal E}_\Delta$}}
\put(-70,30){\makebox(0,0){$E_\Delta$}}
\put(-10,30){\makebox(0,0){${E}^{st}_\Delta$}}
\put(-40,0){\makebox(0,0){${\cal E}^{st}_\Delta$}}

\put(-45,55){\vector(-1,-1){20}}
\put(-25,15){\vector(1,1){10}}
\put(-20,20){\vector(-1,-1){13}}
\put(-35,55){\vector(1,-1){20}}

\put(90,60){\makebox(0,0){${\cal D}_T$}}
\put(60,30){\makebox(0,0){$D_T$}}
\put(120,30){\makebox(0,0){${D}^{st}_T$}}
\put(90,0){\makebox(0,0){${\cal D}^{st}_T$}}

\put(85,55){\vector(-1,-1){20}}
\put(105,15){\vector(1,1){10}}
\put(110,20){\vector(-1,-1){13}}
\put(95,55){\vector(1,-1){20}}

\thicklines
\put(10,30){\vector(1,0){30}}

\put(20,-25){\makebox(0,0){$\frac{\alpha:\beta}{\gamma}$\ \
$\Rightarrow$\ \ {\scriptsize $K\alpha \wedge \neg K\neg \beta \supset
\gamma$}}}
\end{picture}
\end{center}
\caption{Embedding default logic into autoepistemic logic}
\label{fig3}
}
\end{figure}

Once we properly align concepts from default logic with those from
autoepistemic logic, Konolige's translation works! This alignment is
illustrated in Figure \ref{fig3} and is formally described in 
the following theorem.

\begin{theorem}
Let $\Delta$ be a default theory and let $T=m(\Delta)$. Then the
following pairs of operators coincide and, thus, have the same
fixpoints:
\begin{enumerate}
\item $E_{\Delta} = D_T$ (that is, weak extensions correspond to
expansions).
\item ${\cal E}_{\Delta}= {\cal D}_T$ (that is, partial weak
extensions correspond to partial expansions; Kripke-Kleene semantics for
$\Delta$ and Kripke-Kleene semantics for $T$ coincide).
\item $E_\Delta^{st}$ and $D_T^{st}$ (that is, extensions correspond to
strong expansions).
\item ${\cal E}^{st}_\Delta$ and ${\cal D}^{st}_T$ (that is,
partial extensions correspond to partial strong expansions,
well-founded semantics for $\Delta$ and well-founded semantics for 
$T$ coincide).
\end{enumerate}
\end{theorem}

\section{DISCUSSION AND FUTURE WORK}

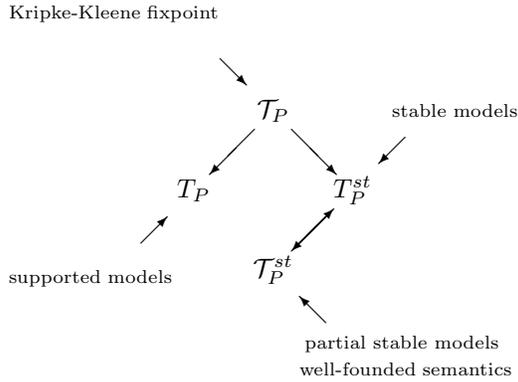
\begin{figure}[t]
\begin{center}
\begin{picture}(60,150)(-30,-40)
\put(-100,-5){\scriptsize supported models}
\put(-50,10){\vector(1,1){10}}
\put(-100,95){\scriptsize Kripke-Kleene fixpoint}
\put(-20,80){\vector(1,-1){10}}
\put(0,60){\makebox(0,0){${\cal T}_P$}}
\put(-30,30){\makebox(0,0){${T}_P$}}
\put(12,-30){\scriptsize partial stable models}
\put(10,-40){\scriptsize well-founded semantics}
\put(50,50){\vector(-1,-1){10}}
\put(45,58){\scriptsize stable models}
\put(20,-20){\vector(-1,1){10}}
\put(30,30){\makebox(0,0){$T^{st}_P$}}
\put(0,0){\makebox(0,0){${\cal T}^{st}_P$}}
\put(-7,53){\vector(-1,-1){17}}
\put(10,10){\vector(1,1){13}}
\put(20,20){\vector(-1,-1){13}}
\put(7,53){\vector(1,-1){17}}
\end{picture}
\end{center}
\caption{Operators associated with logic programming}
\label{fig4}
\end{figure}

We presented results uncovering the semantic properties of default
and autoepistemic logics. In each case, a whole family of semantics can
be derived from a single operator by purely algebraic transformations.
Most importantly, the translation of Konolige establishes a perfect
correspondence between the families of semantics of default and
autoepistemic logics. This elegant picture can be further extended
to the case of logic programming. As discovered by Fitting, all key
semantics for logic programs can be similarly obtained from a single
operator, the 4-valued van Emden-Kowalski one-step provability operator
\cite{vEK76}. The resulting semantic structure for logic programming is 
shown in Figure \ref{fig4}. 

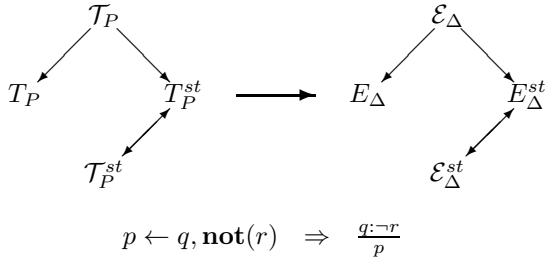
\begin{figure}[t]
\begin{center}
\begin{picture}(60,105)(0,-30)
\put(-40,60){\makebox(0,0){${\cal T}_P$}}
\put(-70,30){\makebox(0,0){$T_P$}}
\put(-10,30){\makebox(0,0){${T}^{st}_P$}}
\put(-40,0){\makebox(0,0){${\cal T}^{st}_P$}}

\put(-45,55){\vector(-1,-1){20}}
\put(-25,15){\vector(1,1){10}}
\put(-20,20){\vector(-1,-1){13}}
\put(-35,55){\vector(1,-1){20}}

\put(90,60){\makebox(0,0){${\cal E}_\Delta$}}
\put(60,30){\makebox(0,0){$E_\Delta$}}
\put(120,30){\makebox(0,0){${E}^{st}_\Delta$}}
\put(90,0){\makebox(0,0){${\cal E}^{st}_\Delta$}}

\put(85,55){\vector(-1,-1){20}}
\put(105,15){\vector(1,1){10}}
\put(110,20){\vector(-1,-1){13}}
\put(95,55){\vector(1,-1){20}}

\thicklines
\put(10,30){\vector(1,0){30}}

\put(20,-25){\makebox(0,0){$p\lar q, \n(r)\ \ \Rightarrow\ \ \frac{q:\neg
r}{p}$}}
\end{picture}
\end{center}
\caption{Embedding logic programming in default logic}
\label{fig5}
\end{figure}

In addition, the translation of logic program clauses into default rules 
proposed in \cite{bf91a,mt89c} establishes an embedding of logic 
programming into default logic that precisely aligns the corresponding 
semantics (Figure \ref{fig5}).

Let us further note that the approach to semantics of nonmonotonic
logics presented here can also be extended to the case of reflexive 
autoepistemic logic by Schwarz \cite{sch91}. As in all other cases
discussed in this paper, all major semantics for the reflexive
autoepistemic logic can be obtained from a single approximating operator
closely related to the operator ${\cal D}_T$.

Our work also points to an interesting open problem in the area of modal
nonmonotonic logics. The fixpoints of the operator $D^{st}_T$ have not, to
the best of our knowledge, been studied in the literature. In particular,
it is not known if they can be described in the McDermott-Doyle scheme
\cite{mcdd80,mcd82} as $\cal S$-expansions for some modal logic $\cal
S$.

Finally, let us mention that it is possible to develop an abstract, purely 
algebraic treatment of the concept of approximations. It generalizes 
the approach presented here and the work of Fitting on logic programming 
semantics. An account of this abstract treatment of approximations can be
found in \cite{dmt00}. 

\subsubsection*{Acknowledgements}

This work was partially supported by the NSF grants CDA-9502645 and
IRI-9619233.

\end{document}